%% file: elsarticle-template-1-num.tex
\documentclass[12pt]{elsarticle}

\makeatletter

\usepackage{graphicx}
\usepackage{amsmath}
\usepackage{url}
\usepackage{multirow}
\usepackage{siunitx}
\usepackage{graphbox}

\newcommand{\simtosim}{\texttt{sim} $\rightarrow$ \texttt{sim}}
\newcommand{\exptoexp}{\texttt{exp} $\rightarrow$ \texttt{exp}}
\newcommand{\simtoexp}{\texttt{sim} $\rightarrow$ \texttt{exp}}

\graphicspath{ {images/} }
\usepackage{amssymb}

\usepackage{lineno}

\journal{Nuclear Instruments and Methods A}

\begin{document}

\begin{frontmatter}


\title{Machine Learning Methods for Track Classification in the AT-TPC}



\author[DavPhy]{M.P. Kuchera}
\author[DavCS]{R. Ramanujan}
\author[DavPhy]{J. Z. Taylor}
\author[DavCS]{R. R. Strauss}
\author[MSU]{D. Bazin}
\author[MSU]{J. Bradt}
\author[DavPhy]{Ruiming Chen}
\address[DavPhy]{Department of Physics, Davidson College, Davidson, NC 28035, USA}
\address[DavCS]{Department of Mathematics and Computer Science, Davidson College, Davidson, NC 28035, USA}
\address[MSU]{National Superconducting Cyclotron Laboratory, Michigan State University, Lansing, Michigan, 48824, USA}

\input{abstract}

\end{frontmatter}


\input{intro}

\input{ml}

\input{fitting}

\input{results}

\input{conclusions}

\input{acknowledgements}





\bibliographystyle{model1-num-names}
\bibliography{els_classification.bib}







\end{document}

%% file: abstract.tex
\begin{abstract}

We evaluate machine learning methods for event classification in the Active-Target Time Projection Chamber detector at the National Superconducting Cyclotron Laboratory (NSCL) at Michigan State University. Currently, events of interest are selected via cuts in the track fitting stage of the analysis workflow. An explicit classification step to single out the desired reaction product would result in more accurate physics results as well as a faster analysis process. We tested binary and multi-class classification methods on data produced by the $^{46}$Ar(p,p) experiment run at the NSCL in September 2015. We found that fine-tuning a pre-trained convolutional neural network produced the most successful classifier of proton scattering events in the experimental data, when trained on both experimental and simulated data. We present results from this investigation and conclude with recommendations for event classification in future experiments.

\end{abstract}


%% file: intro.tex
\section{Introduction}
\label{Intro}

\subsection{Challenges of data analysis in the Active-Target Time Projection Chamber}
\label{sec:attpc-challenges}
A time projection chamber (TPC) is a particle detector that is capable of full three-dimensional reconstruction of charged particles traveling through the detector medium. The Active-Target Time Projection Chamber (AT-TPC) is a TPC filled with a gas that acts as both the target and the detection medium for nuclear reactions. This detector is used for low energy nuclear physics experiments that study exotic nuclei at the National Superconducting Cyclotron Laboratory (NSCL) at Michigan State University \cite{Bradt-ATTPC}. \\

A detector providing high-resolution, three-dimensional tracks with nearly $4\pi$ resolution is an ideal setup for experiments that have low reaction rates. However, a typical week-long experiment using the AT-TPC generates on the order of 10 terabytes of raw data, from which charge deposition and spatial data is extracted. Track fitting to extract energy and angular information, and event classification for particle identification are challenging problems in this setting. This work proposes to decouple the fitting and classification stages of the analysis, which are described in Sec.~\ref{sec:ATTPC} and in detail in \cite{Bradt-ATTPC}. We propose the use of machine learning methods to classify events in the AT-TPC.\\

Other TPCs, such as MicroBooNE, a liquid argon TPC, have successfully applied machine learning methods to classify particle tracks \cite{Acciarri_2017}. The data that MicroBooNE produces can be directly represented as images. Therefore, Convolutional Neural Networks (CNNs) are a natural fit for their learning problem \cite{Acciarri_2017}. \citeauthor{Acciarri_2017} used preexisting CNN network architectures for classifying simulated data, but did not consider the problem of classifying experimental data (or using simulated data to classify experimental data). More recently, MicroBooNE considered the problem of classifying experimental data from models trained on simulated data \cite{Adams:2018bvi}. In addition, CNNs have been shown to be successful in extracting energy and position of events in liquid xenon TPCs \cite{Delaquis_2018}. Building on this work, we present results which fine tune pre-trained models, therefore decreasing training time significantly. In addition, we present results on models that train on simulated data and yet classify experimental data.

\subsection{AT-TPC details}
The AT-TPC detector volume is cylindrical, with a length of $1$m along the beam axis and a radius of $0.292$m. Charged particles are ionized in the detector volume and the free electrons drift ``downstream'' towards the rear end of the detector, where a micromegas device detects the electrons \cite{GIOMATARIS199629}. This detection plane is composed of $10,240$ triangular electrodes, or {\em pads}, which define the position resolution of the signal in the plane perpendicular to the beam axis \cite{GIOMATARIS199629}. In the $^{46}$Ar(p,p) experiment, $512$ time steps were recorded. This generates $5,242,880$ voxels per event. Each voxel is represented by a floating point number.\\

The AT-TPC is designed for low-energy nuclear physics experiments with low reaction rates, which means high efficiency is required for collecting sufficient statistics. The AT-TPC allows for nearly $4\pi$ angular coverage and is capable of detecting all triggered events in the detector volume. In addition, the gas in the chamber also acts as the target, allowing for detection of reactions at a continuous range of beam energies. This allows for the construction of excitation functions over a wide range of energies, as seen in \cite{BRADT2018155}. For a full description of the detector, we refer the reader to \cite{Bradt-ATTPC}.

\subsection{Event classification in the AT-TPC}
\label{sec:ATTPC}

\begin{centering}
\begin{figure}[htb]
\includegraphics[width=0.9\textwidth]{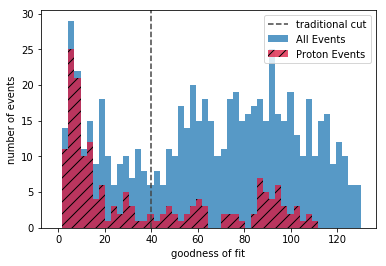}
\caption{Histogram visualizing the classification cut for hand-labeled events from Run 130 in the traditional analysis. Based on the ``goodness of fit" distribution of all data, a cut (dashed line) was chosen at 40 (in arbitrary units). The events that were hand-classified as protons from this run are hatched.}
\label{fig:traditional}
\end{figure}
\end{centering}

The traditional event classification procedure is described in detail in \cite{Bradt-ATTPC}. The steps are outlined below.
\begin{enumerate}
\item Clean all events using Hough transform methods. Discard events that have fewer than some preset number of points.
\item Use the na\"ive Monte Carlo method to fit every track, assuming that each represents the desired reaction product. The best fit minimizes the objective function defined in \cite{Bradt-ATTPC}.
\item Plot a histogram of the objective function for all events.
\item Visually determine a cutoff for a ``good fit", or low objective function value. This cutoff for our data is shown in Fig.~\ref{fig:traditional}
\item Define all events within that cut as events of interest.
\end{enumerate}
The analysis workflow currently has no step which can verify that the events accepted in the cut are from the desired reaction. Further, the method does not quantify how many desired events were eliminated by this cut. The goal of this work is to decouple the classification and fitting steps in the analysis workflow. In addition, we aim to classify events as early in the analysis process as possible in order to remove potential bias from subsequent steps. For example, the cleaning algorithm (step 1 above) expects curved tracks with a well-defined center of curvature. However, when presented with tracks without a spiral signature --- caused, for example, by electronic sparks or cosmic ray interference --- this algorithm still treats them as such, and uses this center to determine which points within an event are noise and which constitute the signal. This degrades the efficacy of any ensuing analysis steps. Therefore, we would like to classify events as early in the analysis workflow as possible.\\

The experiment tested in this work had a beam of $^{46}$Ar nuclei incident on isobutane gas with the goal of studying proton elastic scattering reactions. The details of the experiment and the results derived using the traditional methods of analysis are presented in \cite{BRADT2018155}. The $^{46}$Ar experiment ran for $12$ days and produced approximately $12$ TB of raw data. After initial cuts on the data to remove events from beam contaminants and pileup, we estimate from this work that approximately $25\%$ of the remaining data comprises proton elastic scattering events, which is the reaction of interest to the researchers. This value was estimated from manually labeling a small subset of the raw data, which is discussed in Sec.~\ref{section:experimental-data} and used in Fig.~\ref{fig:traditional}. Our goal for this work was to create a model to automatically select these proton scattering events from the spatial and charge information for each event.

%% file: ml.tex
\section{Machine Learning Background}
\label{sec:ml}
 
\begin{figure}[htb]
    \centering
    \includegraphics[width=0.5\textwidth]{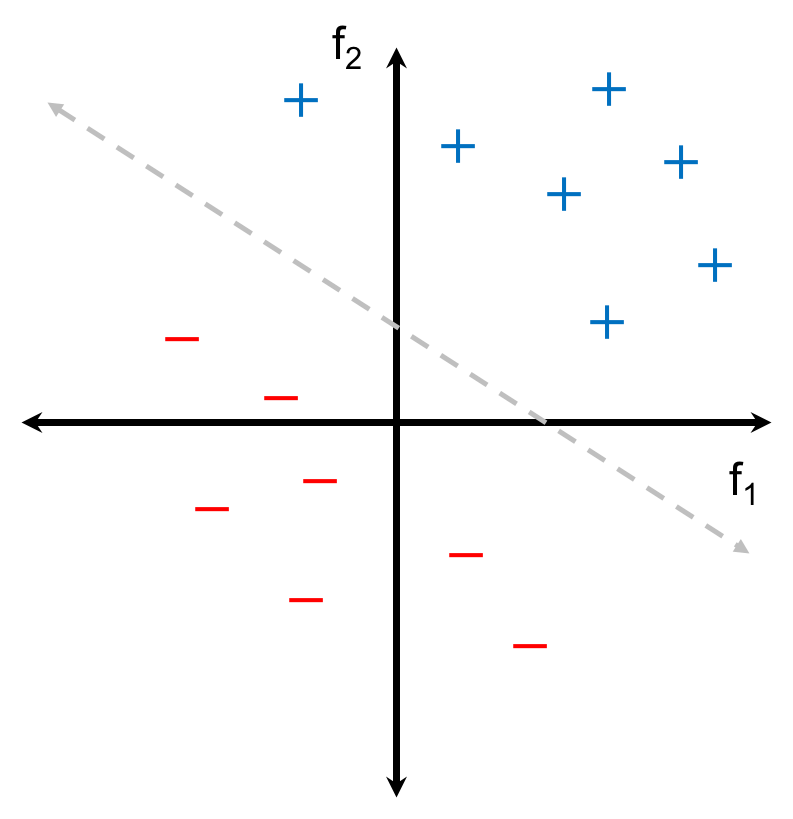}
    \caption{A separating hyperplane for a classification problem where each example is represented by two features $f_1$ and $f_2$. While the two classes are cleanly separated by a line in this example, more complicated data distributions may require the use of non-linear surfaces.}
    \label{fig:separating-hyperplane}
\end{figure}

The current approach for identifying events of interest from the experimental data relies on Monte-Carlo optimization techniques in conjunction with a $\chi^2$-based goodness-of-fit test, as discussed in Sec.~\ref{sec:ATTPC}. In this work, we propose using machine learning methods instead for the problem of track classification in the AT-TPC. We start with a dataset comprising descriptions of events recorded in the detector. Each event is represented as a vector of real numbers (of fixed length) and is termed an \emph{example}. Each component of this vector is called a \emph{feature}. Attached to every example is a \emph{label}, indicating the identity of the particle that generated the said track --- for example, proton or non-proton. The learning task is to find a surface $h$ (dubbed a \emph{separating hyperplane}) that partitions the data such that all tracks corresponding to proton events lie on one side of $h$, while all non-proton events fall on the other side. Formally, we seek a function $h_{\theta}: \mathbb{R}^n \mapsto \{0, 1\}$, parameterized by a set of weights $\{\theta_1, \theta_2, \ldots, \theta_n\}$, that minimizes a \emph{cost function} $J(\boldsymbol{\theta})$. The function $J$ is a cumulative measure of the classification errors made by the decision surface $h_{\theta}$. Fig.~\ref{fig:separating-hyperplane} provides the visual intuition behind this task, using a simplified scenario where every event is represented by a two-dimensional point and a line is an adequate decision boundary. In practice, however, the events are usually not easily separable; thus, we also investigate the effectiveness of non-linear separating surfaces for the track classification problem\footnote{Though it should be noted that perfect separability is seldom desirable as it suggests that \emph{overfitting} has occurred, which is discussed further in Sec.~\ref{sec:overfitting}.}. Appropriate values of the weight parameters $\boldsymbol{\theta}$ are found using standard optimization techniques like gradient descent. After preliminary explorations, we focused our study on three model families, which are described below.

\subsection{Logistic regression}

Logistic regression is a classic approach in statistics and machine learning for modeling the probable value of a binary variable $y$. If we (arbitrarily) assign proton tracks the label $y=1$ and non-proton tracks the label $y=0$, then given an example $\boldsymbol{x} \in \mathbb{R}^n$ representing an event, the logistic regression model calculates the probability that this was a proton event as:
$$ Pr(y = 1 | \boldsymbol{x}) = h_{\theta}(\boldsymbol{x}) = \frac{1}{1 + e^{-\boldsymbol{\theta}^T \cdot \boldsymbol{x}}} $$
The expression $\boldsymbol{\theta}^T \cdot \boldsymbol{x}$ represents a weighted linear combination of the input features. This operation is composed with a \emph{mean function} --- in this case, the logistic function $\sigma(z) = 1/(1 + e^{-z})$ --- to ensure that the output of the model remains bounded by $[0, 1]$. The parameters $\boldsymbol{\theta}$ are fitted by minimizing the binary cross-entropy loss function:
\begin{equation} \label{eq:cost}
J(\boldsymbol{\theta}) = -\frac{1}{m}\sum_{i=1}^{m} \left( y_i \log{\hat{y}_i} + (1-y_i)\log{(1-\hat{y}_i)} \right)
\end{equation}
where $m$ denotes the number of examples in the dataset and $\hat{y}_i = h_{\theta}(\boldsymbol{x}_i)$ is the model's prediction for the $i^{th}$ event $\boldsymbol{x}_i$ \cite{HTF}. Smaller values of this loss function $J$ are achieved when the model's predictions $\hat{y}$ more closely match the true labels $y$. While the treatment above considers the scenario of binary classification, where each example belongs to one of two categories, logistic regression can be extended in a straightforward fashion to the multi-class setting as well --- we refer the reader to \cite{HTF} for further details.

\subsection{Neural networks}

The term \emph{neural network} describes a large class of models that are loosely inspired by the processes that drive learning in biological neurons. They have been successfully applied to problems from a number of challenging domains, including image recognition \cite{inceptionv4}, autonomous driving \cite{deepdriving}, game-playing \cite{alphago}, as well as in experimental physics \cite{DENBY}. A neural network comprises a collection of units (``neurons'') that form a graph. Each unit accepts inputs from other units, aggregates them in some non-linear manner and produces an output that may be consumed by other units. This composition of neuronal units means that such networks are able to learn very complex decision surfaces; indeed, under some mild assumptions, it can be shown that neural networks with a finite number of units can approximate any continuous function to arbitrary precision \cite{nn-universality}. While neural networks are highly configurable, only certain network architectures and neuronal unit choices find wide use in practice. In the next two subsections, we describe the two neural network models that were explored in this work.

\subsubsection{Fully-connected feed-forward neural networks}
\label{sec:fully-connected}

\begin{figure}[htb]
    \centering
    \includegraphics[align=c,width=0.4\textwidth]{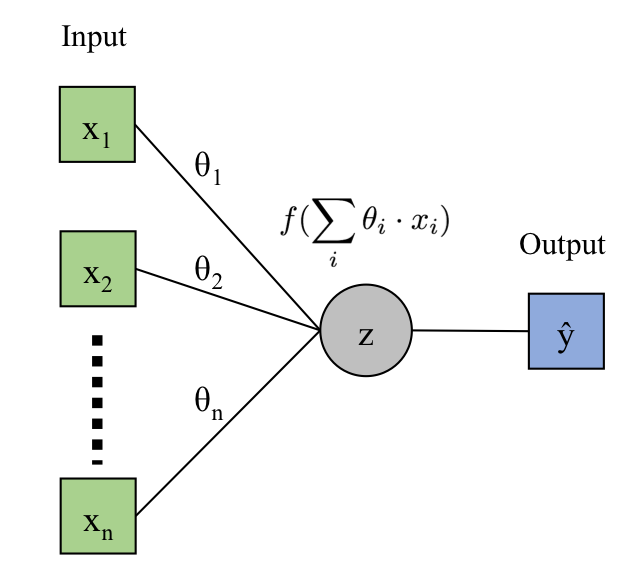}
    \hspace{0.3in}
    \includegraphics[align=c,width=0.5\textwidth]{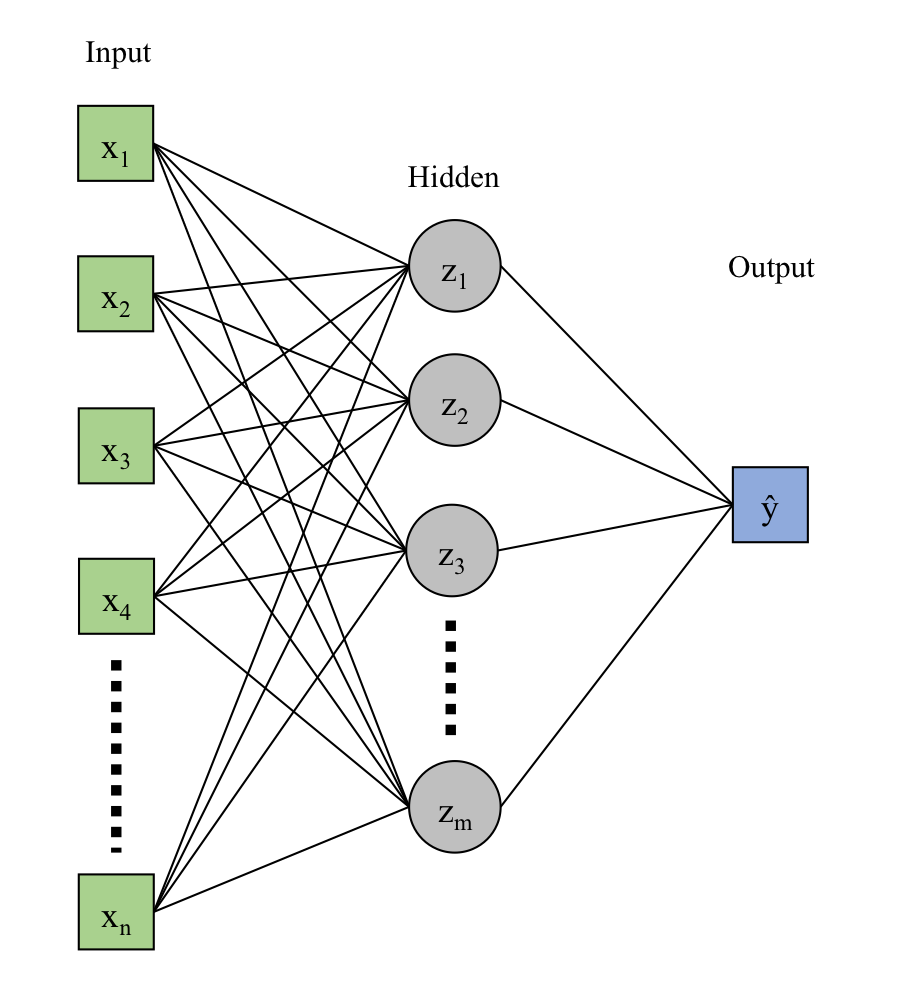}
    \caption{Left: A single neuronal unit. The function $f$ is a non-linear activation function. Right: A fully-connected feed-forward neural network formed by connecting together several artificial neurons.}
    \label{fig:ann}
\end{figure}

In this classic neural network architecture, every node computes a linearly weighted sum of its inputs (plus a bias term). A non-linear \emph{activation function} is then applied to this sum to produce the unit's output. We use the rectified linear function ($f(z) = \max(0, z)$) in this work, as it produced superior results to other common choices like the hyperbolic tangent and logistic functions. The left panel in Fig.~\ref{fig:ann} visually depicts the computation performed by each artificial neuron.\\

In a \emph{fully-connected} architecture, the individual neurons are organized in a layered fashion as depicted in the right panel of Fig.~\ref{fig:ann}. Every unit in a given layer $i$ accepts inputs from every neuron in layer $i-1$ and each connection is associated with its own weight. The term \emph{feed-forward} refers to the fact that the network is acyclic: information flows from the features $\boldsymbol{x}$ that comprise the \emph{input layer}, through one or more \emph{hidden layers} of neurons, to produce a result at the \emph{output layer}, with no feedback loops. Networks with more than one hidden layer are called \emph{deep networks}, and the term \emph{deep learning} refers to the set of techniques used to train such networks \cite{deeplearning}. The overall output of the model $\hat{y} = h_{\theta}(\boldsymbol{x})$ is a composition of applications of the activation function to combinations of the input features $x_i$ and is the estimated probability that the input example corresponds to a proton event $Pr(y = 1 | \boldsymbol{x})$. As with logistic regression, the parameters of the model (i.e., the weights attached to each edge in the network) are fitted by minimizing the cross-entropy loss defined by Eq.~\ref{eq:cost} using optimization techniques like gradient descent. The \emph{backpropagation} algorithm offers a method for efficiently computing the gradient of the loss function with respect to the network weights by caching the results of intermediate derivatives and avoiding their repeated recomputation \cite{backprop}.

\subsubsection{Convolutional neural networks}
\label{sec:cnn}

A \emph{convolutional neural network} (CNN) is a learning model that is particularly suited for applications that operate on grid-like data, such as time series (a $1$-dimensional grid of samples) or images (a $2$-dimensional grid of pixels). While their early success in problems like recognizing hand-written zip codes \cite{cnn} hinted at their potential, CNNs truly came into their own in $2012$ when the \texttt{AlexNet} model resoundingly won the ImageNet visual recognition contest \cite{alexnet}. CNNs have since completely revolutionized the field of computer vision, and are central to state-of-the-art methods for many challenging problems such as object recognition \cite{inceptionv4}, image segmentation \cite{image-segmentation}, and image captioning \cite{image-captioning}. More pertinently, CNNs have also proven to be effective at certain tasks in high-energy physics --- for example, \citeauthor{cnn-neutrino} successfully used CNNs to recognize neutrino events in the NOvA experiment at Fermilab \cite{cnn-neutrino}, albeit on simulated data. We also already noted the work of the MicroBooNE collaboration in this area in Sec.~\ref{sec:attpc-challenges}. In this work, we study the efficacy of CNNs in the AT-TPC track classification domain by framing the problem as a visual recognition task. We train the network to make predictions based on plots of two-dimensional projections of events recorded in the detector. \\

We now provide a sketch of the main ideas underlying CNNs and refer the reader to \cite{deeplearning} for a more comprehensive treatment. A CNN model is built by arranging three kinds of layers, described below, in various configurations.

\begin{figure}[thb]
    \centering
    \includegraphics[width=0.9\textwidth]{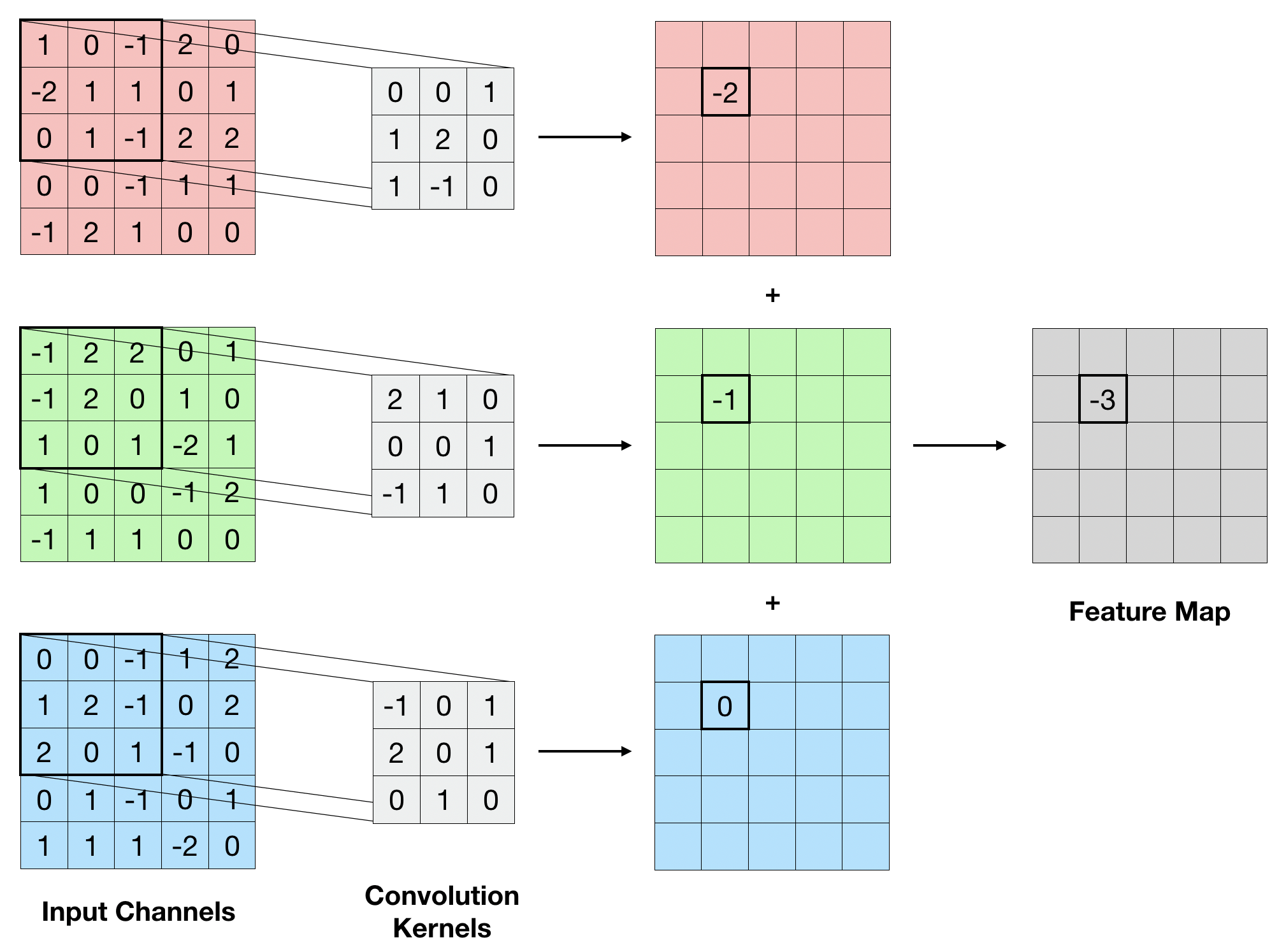}
    \caption{The first step in the convolution of a $5 \times 5 \times 3$ image by a single $3 \times 3$ filter. For each channel of the input, we compute a sum of the highlighted pixel values weighted by the corresponding values in the convolution kernel. The channel-wise results are then summed up to produce the indicated entry in the output feature map. The remaining values of the feature map are populated by sliding the kernel over the input and repeating the computation. Zero-padding is utilized along the edges of the input to ensure that the operation remains well-defined even when parts of the kernel are out of bounds. A convolutional layer comprises many such filters, each of which produces its own feature map. The weights of the kernels are determined by the learning process.}
    \label{fig:convolution}
\end{figure}

\begin{description}
    \item[Convolutional layer] CNNs use the discrete convolution operation to identify distinctive visual characteristics in an image, like corners, lines and shapes. This is a linear transformation that maps a $3$-dimensional input volume to a $3$-dimensional output volume. Formally, consider an input volume (for example, an image) $\boldsymbol{I}$ of dimensions $h \times w \times c$, where $h$ represents the height of the volume, $w$ represents the width, and $c$ the number of channels (for RGB images, $c=3$). An $m \times n \times s$ convolutional layer $\boldsymbol{F}$ comprises a stack of $s$ 2D convolution kernels (also called \emph{filters}), each of which has height $m$ and width $n$. Each filter also extends fully through the channel dimension $c$, and the values produced by the channel-wise convolutions are simply summed up along this axis. The result of convolving $F$ by $I$ is given by: 
    $$ (\boldsymbol{F} * \boldsymbol{I})(i, j, k) = \sum_{p=-m'}^{m'} \sum_{q=-n'}^{n'} \sum_{r=1}^{c} \boldsymbol{I}(i+p, j+q, r) \cdot \boldsymbol{F}(p, q, r, k) $$
    where $(\boldsymbol{F} * \boldsymbol{I})(i, j, k)$ refers to the value at position $(i, j)$ in the $k^{th}$ channel of the output volume, $m' = \left\lfloor \frac{m}{2} \right\rfloor$, and $n' = \left\lfloor \frac{n}{2} \right\rfloor$. Fig.~\ref{fig:convolution} presents this operation visually. Typically, the spatial extent of each kernel is much smaller than the dimensions of the input volume, so that $m \ll h$ and $n \ll w$. Thus, each kernel is highly localized in its sensitivity, and models the notion of a \emph{receptive field} that is a feature of the mammalian visual cortex \cite{receptive-field}. The borders of the input volume are usually padded with an appropriate number of zeros to ensure that the output volume has the same width and height as the input. The output thus has dimensions $h \times w \times s$, and each slice of this volume along the $s$ dimension is referred to as a \emph{feature map}.

    \item[Activation layer] An activation layer typically operates on the output of a convolutional layer and applies a non-linear function to its input in an element-wise fashion. The rectified linear unit (ReLU), defined as $f(x) = \max(0, x)$, is a commonly used activation function that permits effective learning in deep CNNs \cite{alexnet,relu}. 
    
    \item[Pooling layer] A convolutional layer produces a number of feature maps, each of which have the same height and width as the input volume. When a CNN contains many such layers, the memory and computational requirements of the model grow very quickly. Pooling layers alleviate this problem by downsampling the input volume along the spatial dimensions. A commonly used pooling operator is \emph{max pooling}, which replaces the values in an $m \times n$ region of the input with a single value corresponding to the maximum of those values \cite{max-pooling}. The degree of overlap between the pooled regions can be adjusted to reduce the loss of information. Pooling also makes the performance of the network robust to small translations of the input \cite{deeplearning}.
\end{description}

\begin{figure}[htb]
    \centering
    \includegraphics[width=\textwidth]{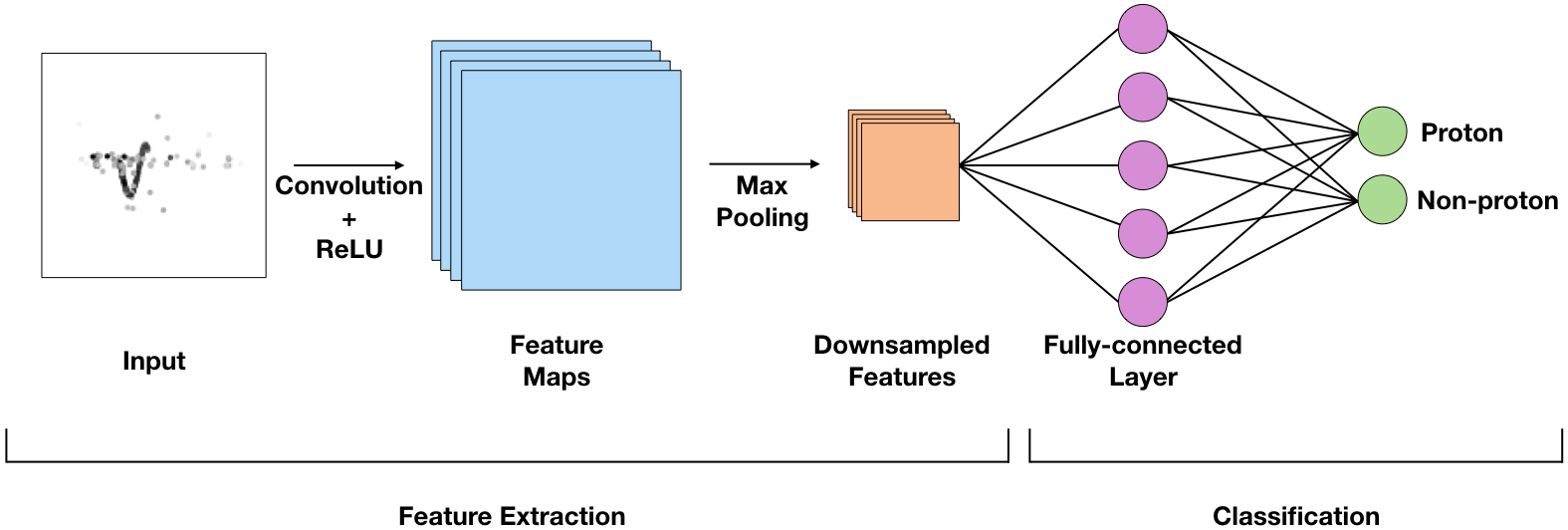}
    \caption{The stages comprising a typical CNN model. In the first stage, the pixel-level data from the input is passed through multiple convolution, activation and pooling layers to extract high-level features describing the image. These features are then used by the latter stage to classify the example.}
    \label{fig:vgg16}
\end{figure}

In addition to the specialized layers described above, a typical CNN also utilizes a fully-connected topology (as seen in Sec.~\ref{sec:fully-connected}) in the last one or two layers preceding the output layer. The role of these dense layers is to combine the features extracted by the prior layers to produce a categorical prediction. A schematic diagram of the overall architecture is shown in Fig.~\ref{fig:vgg16}. As before, the network is trained by minimizing the cross-entropy loss function given by Eq.~\ref{eq:cost}. The learning process tunes the weights associated with the fully connected units, as well as the weights of the convolution kernels.\\

High-performing CNN models are typically composed of many layers. For example, the \texttt{VGG16} model that won the ImageNet visual recognition challenge in 2014 uses $16$ weight layers --- $13$ convolutional layers (combined with ReLU activations) and $3$ fully-connected layers, with max-pooling layers inserted at various stages \cite{vgg16}. The model has {\small $\sim$}$138$ million parameters. Training such a network from scratch, with random initialization, requires extremely large labeled datasets (ImageNet has {\small $\sim$}$14$ million examples \cite{imagenet}) and can take hours to days on high-performance Graphics Processing Units (GPUs). For applications with limited training data, an alternative is to use a \emph{pre-trained network} \cite{feature-transfer} (like \texttt{VGG16} trained on ImageNet) in one of two ways:
\begin{itemize}
    \item For feature extraction: One could remove the final fully-connected and output layers from the pre-trained model and process each training image (for example, two-dimensional plots of AT-TPC events) by passing it through the network. This produces a fixed-length vector representation of the original image, that can be then be used as the input to a simple classification algorithm, like logistic regression.
    \item For fine-tuning: One could use the weight settings of the pre-trained model as a starting point, and adjust these using gradient descent approaches to improve the model's performance on the new dataset. A very small learning rate is used to ensure that the original weights are not subject to excessive distortions.
\end{itemize}
These are examples of \emph{transfer learning}, and we explore both these approaches in this work.

%% file: fitting.tex
\section{Model Fitting Methodology}

There are a number of practical considerations when one builds machine learning models for a particular application. Learning algorithms often have a number of \emph{hyperparameters} --- free variables whose values must be fixed prior to commencing the model fitting process --- whose settings can have a large impact on the performance of the model. These hyperparameters must thus be chosen thoughtfully. Further, care must be taken when measuring the performance of the model, to ensure that the reported results are unbiased and not overly optimistic. This section provides an overview of our model fitting methodology.

\subsection{Combating Overfitting}
\label{sec:overfitting}

\begin{figure}[htb]
\begin{center}
\includegraphics[width=0.95\textwidth]{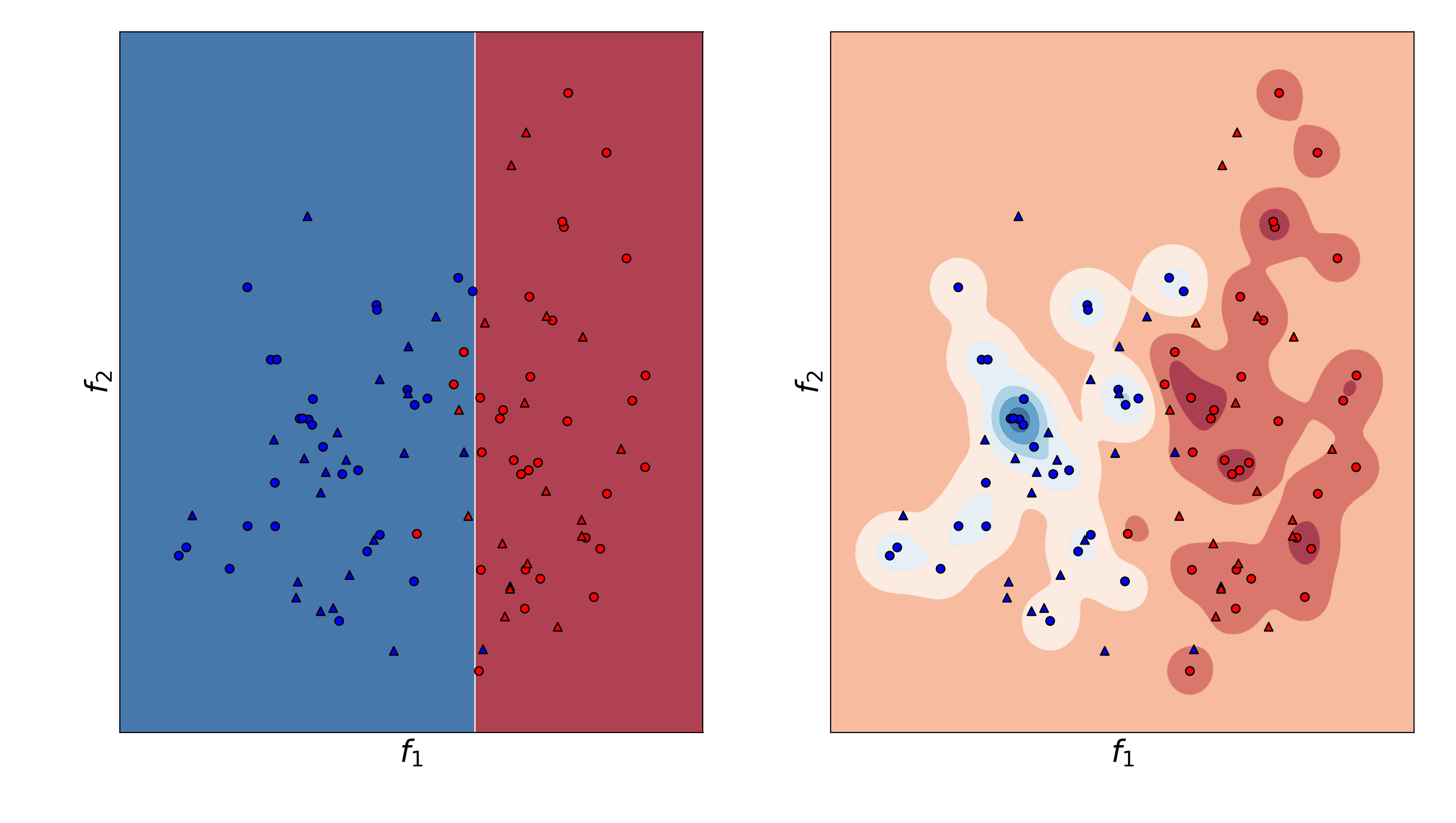}
\caption{The above are scatter plots of an artificial dataset with two features $f_1$ and $f_2$. This is a binary classification task, with red and blue corresponding to the two class labels. The circles are points used for training the model; the triangles are additional data points that are not used for training purposes but to evaluate the learned model's ability to generalize. The shaded regions and their boundaries represent the fitted decision surface. In the left panel, we have a linear model that is well-fit: while it misclassifies some of the training data, it also makes few errors on the unseen data points. The right panel depicts an overfit model, i.e., one that correctly classifies every training example, but makes many mistakes on new, unseen examples.}
\label{fig:overfitting}
\end{center}
\end{figure}

In statistics, the term \emph{overfitting} refers to the situation where a model captures the regularities in the training dataset so well that it fails to generalize effectively to unseen examples \cite{HTF}. This problem is particularly acute when a model has high \emph{capacity} (i.e., the ability to represent complex decision boundaries) or has a large number of parameters relative to the size of the training data. Fig.~\ref{fig:overfitting} provides a visual illustration of overfitting. In the remainder of this section, we describe some techniques that were used to address overfitting in our work.

\subsubsection{Train-Test-Validation Splits}
One common method for estimating the \emph{generalization error} of a learned model is to evaluate its performance on a set of held out data, that was not used during the training process. The original collection of examples is divided into a \emph{training}, a \emph{validation} and a \emph{test} set. The data in the training set is used for the fitting process. The validation set is used to compare different hyperparameter settings (for example, the size and number of hidden units in a neural network) and choose the best performing configuration. The test set is used to report the performance of the final model, and offers an unbiased measure of the model's quality. We follow this protocol in this work: all model performance figures in Sec.~\ref{sec:results} are reported on a held out test set.

\subsubsection{Early Stopping}
\begin{figure}[htb]
\begin{center}
\includegraphics[width=0.85\textwidth]{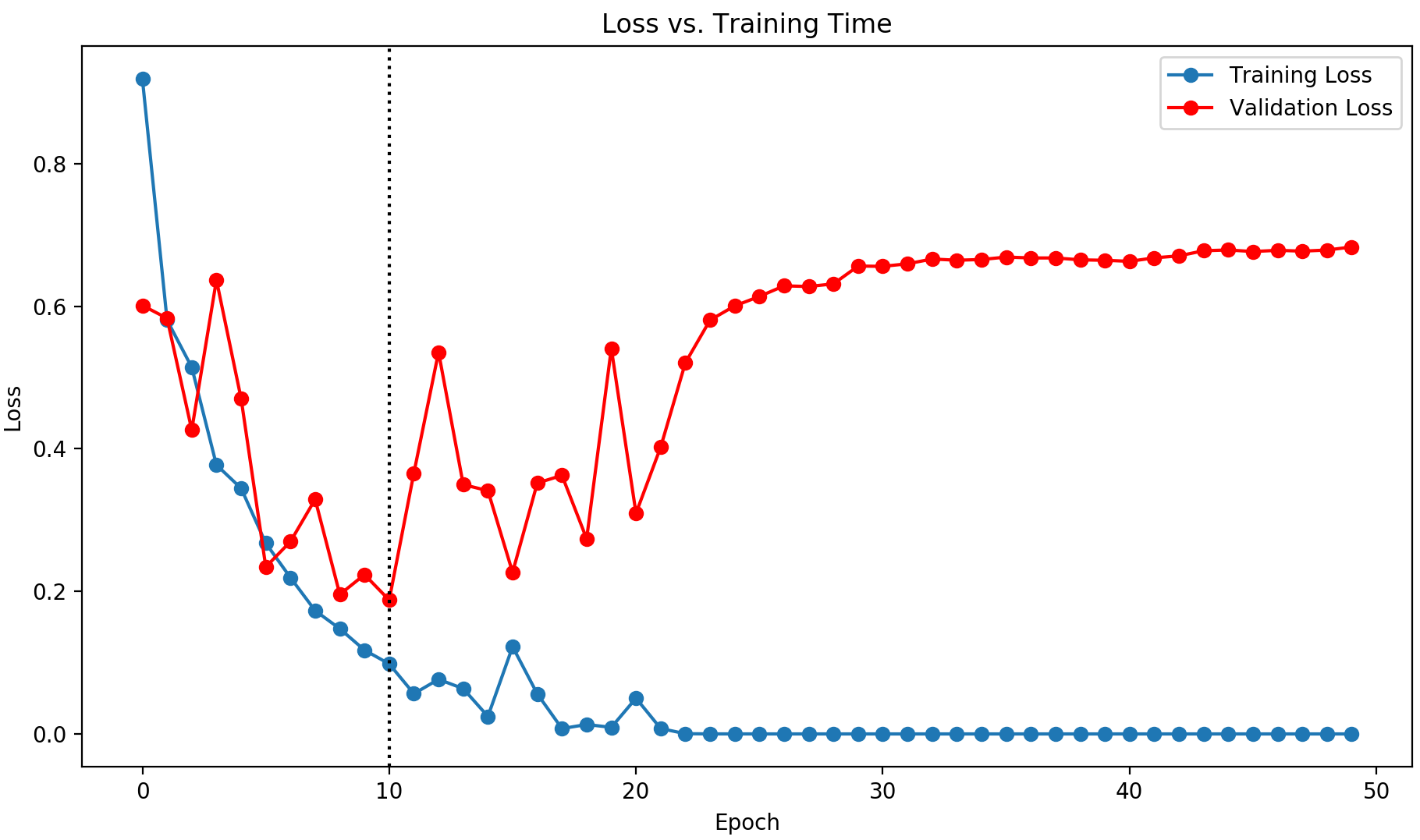}
\caption{A learning curve depicting the value of the model's loss on examples in the training set (blue) and the validation set (red) over time. The $x$-axis indicates \emph{epochs} of training, where one epoch represents one complete pass through the dataset for stochastic gradient descent. The early stopping point in this example occurs at epoch $10$, indicated by the vertical black line.}
\label{fig:learning-curve-loss}
\end{center}
\end{figure}

For models that are fitted using an iterative procedure --- for example, neural networks trained with stochastic gradient descent --- one can mitigate the effects of overfitting by monitoring the progress of the learning process. In particular, one can plot a \emph{learning curve} like that shown in Fig.~\ref{fig:learning-curve-loss}, that tracks the value of the model's loss on the training and validation sets over time. The fitting process can then be stopped at the point where the two losses start diverging, which signals the onset of overfitting.

\subsubsection{Regularization}
\emph{Regularization} is a mathematical technique that discourages the fitting of overly complex models by including a penalty term in the loss function being minimized by the training process \cite{HTF}. The squared L2 norm of the model's parameter vector $\boldsymbol{\theta}$ is a common choice for the regularizer; when combined with Eq.~\ref{eq:cost}, it yields the following modified binary cross-entropy loss function:

\begin{equation*}
J(\boldsymbol{\theta}) = -\frac{1}{m}\sum_{i=1}^{m} \left( y_i \log{\hat{y}_i} + (1-y_i)\log{(1-\hat{y}_i)} \right) + \frac{\lambda}{2m} \sum_{j=1}^{n} \theta_j^2
\end{equation*}

Here, $\lambda \geq 0$ is a hyperparameter that controls the strength of the regularization. Larger values of $\lambda$ cause the weights in the model to shrink and yield simpler decision boundaries \cite{HTF}. The ideal setting of $\lambda$ for a given problem can be found by sweeping a range of values and evaluating the performance of each resulting model on the validation set.

\subsubsection{Dropout}
\emph{Dropout} is a regularization technique that is widely used when training deep neural networks \cite{dropout}. At training time, before each batch of examples is presented to the network, a random subset of the neuronal units in the model (along with their connections) is temporarily removed. The parameters of the resulting pruned model are adjusted as per the usual training procedure. A different thinned model is then sampled before the next batch of examples, and this process is repeated until the end of the training phase. At test time, the trained connection weights in the model are scaled by the dropout probability $p$, which is a hyperparameter. This randomized dropping of units during training has been shown to clearly benefit performance \cite{dropout} and is thus used by several well-known deep models such as \texttt{AlexNet} \cite{alexnet} and \texttt{VGG16} \cite{vgg16}.

\subsection{Performance Metrics}

In this work, we measure the performance of our machine learning models using three metrics that are widely used by the information retrieval and machine learning communities: \emph{precision}, \emph{recall}, and \emph{F1 score} \cite{IR}. Compared to more intuitive measures like classification accuracy, these metrics are more effective at conveying the nuances of a model's performance and provide insight into the nature of its errors. In the AT-TPC track classification domain, we are particularly interested in a model's ability to correctly identify proton events, and these metrics enable us to characterize this precisely.

\begin{figure}[htb]
\begin{center}
\includegraphics[width=0.85\textwidth]{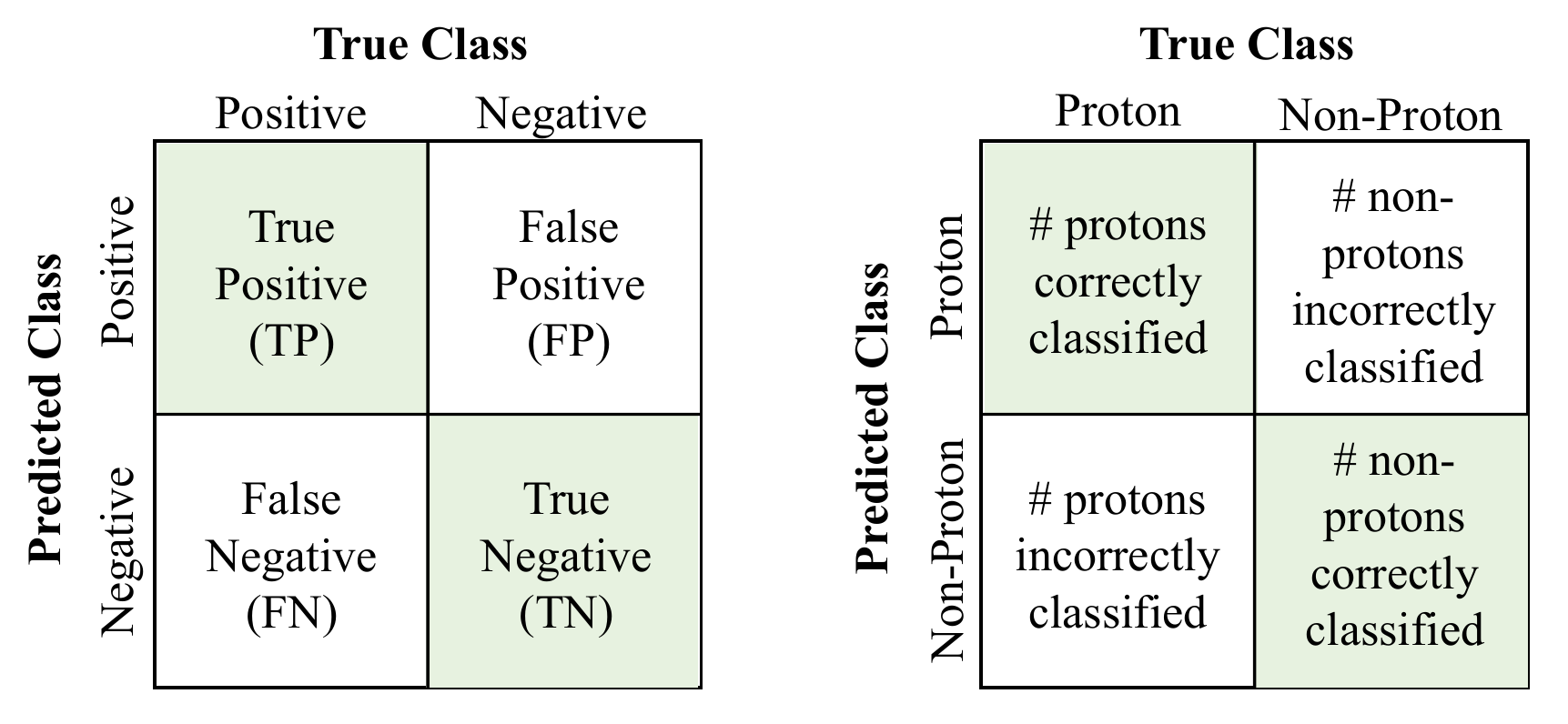}
\caption{The elements of a confusion matrix for a binary classification problem. Each row represents the class predicted by the classifier, while each column represents the true class label. The matrix on the left presents the terms used to describe each entry. The matrix on the right interprets these terms when applied to AT-TPC data. A perfect classifier induces a diagonal confusion matrix.}
\label{fig:confusion-matrix}
\end{center}
\end{figure}

The definitions of precision and recall are best understood by first examining a structure known as a \textit{confusion matrix}. A confusion matrix tabulates the predictions made by a binary classifier and places them in one of four cells, as shown in Fig.~\ref{fig:confusion-matrix}. This figure also presents the terms used to refer to the entries in a confusion matrix and a contextual example using AT-TPC data categories. Precision is a measure of a classifier's soundness --- in the AT-TPC context, it quantifies how often a model is correct when a track is identified as having been generated by a proton. Using the terminology of Fig.~\ref{fig:confusion-matrix}, we can define precision formally as follows:
\begin{equation*}
\text{precision} = \frac{\text{TP}}{\text{TP} + \text{FP}}
\end{equation*}
Recall, on the other hand, is a measure of a classifier's completeness --- it measures how many proton events were successfully picked out by the model from the set of all true proton events. Formally, the recall is defined as:
\begin{equation*}
\text{recall} = \frac{\text{TP}}{\text{TP} + \text{FN}}
\end{equation*}
\vspace{0.05in}

In most applications, higher precision often comes at the cost of lower recall, and vice versa. Thus, these two metrics are usually measured and reported together. Alternately, their harmonic mean is reported as a single value known as the F1 score: 
\begin{equation*}
\text{F1} = 2 \cdot \frac{\text{precision} \cdot \text{recall}}{\text{precision} + \text{recall}}
\label{equ:F1}
\end{equation*}
All three metrics are bounded by $[0, 1]$, with a higher value indicating better performance on that measure. The results presented in Sec.~\ref{sec:results} report the precision, recall and F1 score of the proton class.

%% file: results.tex
\section{Applications to $^{46}Ar(p, p)$}

To test our methods, we attempt to classify protons against all other reaction products and other charged particles detected in the AT-TPC during the $^{46}Ar(p, p)$ experiment (e13306b) at the NSCL. Ideally, the desired reaction product will be classified early in the analysis workflow. This would occur either in real-time as the experiment runs, or shortly after, on minimally processed data. Realizing this goal requires access to a trained classifier before the experimental data becomes available. One way to achieve this is to fit a model using simulated data instead. 
To investigate the feasibility of this goal, we focus on three machine learning tasks:
\begin{itemize}
    \item training and testing models on simulated data (henceforth, denoted as \simtosim),
    \item training and testing models on real experimental data (denoted \exptoexp), and
    \item training models on simulated data, but testing them on experimental data (denoted \simtoexp).
\end{itemize}
While attaining good performance on simulated data does not achieve our goal of effectively classifying experimental data, this setting provides us with an upper-bound on the performance that can be achieved by our models. \\

In each of the above settings, we also consider two different framings of the learning task: as a binary classification problem, where the model makes a determination of whether an event is a proton scattering event or not, and as a multi-class problem, where the model categorizes an event as one of ``proton'', ``carbon'', or ``other''. The latter approach is motivated by the fact that carbon atoms are a very common by-product in the $^{46}Ar(p, p)$ experiment and the physics of their behavior in the detector is simulable. This allows us to generate synthetic datasets that more closely match the experimental data. We also investigate the question of whether having more fine-grained distinctions among events improves the performance of our models. The ``other'' label is a catch-all category that covers event types that we cannot simulate accurately --- we resort to simple uniform random noise to mimic these events in our synthetic data.

\subsection{Experimental training data} \label{section:experimental-data}

We used data collected from the experiment described in Sec.~\ref{sec:ATTPC}. Supervised machine learning requires labeled data for both training and evaluating the performance of a model. Therefore, a sample of events from the $^{46}Ar(p,p)$ experiment was manually labeled for model building purposes. The proton events are easy to visually identify in a strong magnetic field ({\small $\sim$}$2$ T), given the difference in charge and mass between protons and our other common reaction byproduct, carbon. This allows for simple and relatively error-free hand labeling. Tab.~\ref{table:datasets} summarizes the number of events in our hand-labeled experimental dataset. The size of this dataset is limited only by human effort in labeling the data, as we have over $10^6$ events from the experiment \cite{Bradt-thesis}.

\begin{table}[h!]
\centering
	\begin{tabular}{c| c |c} 
 	Event Type & Simulated & Experimental  \\
 	\hline
 	Proton         & 28000 & 663   \\ 
 	Carbon        & 28000 & 340  \\
 	Other           & 28000 & 1686  \\ 
\hline
	\textbf{Total} & 84000 & 2689 \\ [1ex] 
	\end{tabular}
	\caption{Labeled dataset sizes}
	\label{table:datasets}
\end{table}

\subsection{Simulated training data}

By simulating data, we can create a labeled dataset of proton and carbon events of any desired size. We sample uniformly from $z$, $\phi$, and $\theta$, as opposed to a more realistic distribution, to increase the proportion of events with less common trajectories in our dataset. The uniform distribution allows for more even representation of all types of trajectories within each class, even those at scattering angles with low cross-sections. Additionally, we filter out all simulated tracks that do not contain at least 150 hits in the pad plane, and whose average distance from $(x,y) = (0,0)$ in the $xy$-plane is less than $135$mm, to ensure that our simulated charged particle events resemble the ones we observe in our experimental data.\\

Using the specifications of the $^{46}Ar(p,p)$ experiment described in Sec.~\ref{sec:ATTPC}, we simulate the ideal detector response to proton and carbon tracks in the detector volume using the \texttt{pytpc} software package \cite{Bradt-ATTPC}. This package uses the Bethe-Bloch and Lorentz force equations, which track the particle through the detector volume, and models electron diffusion to produce a pad plane projection with full charge trace information. The simulation produces data in the same format as the real experimental data. Tab.~\ref{table:datasets} presents the size of the simulated dataset used in this work.\\

\begin{figure}[ht]
\begin{center}
\includegraphics[width=0.8\textwidth]{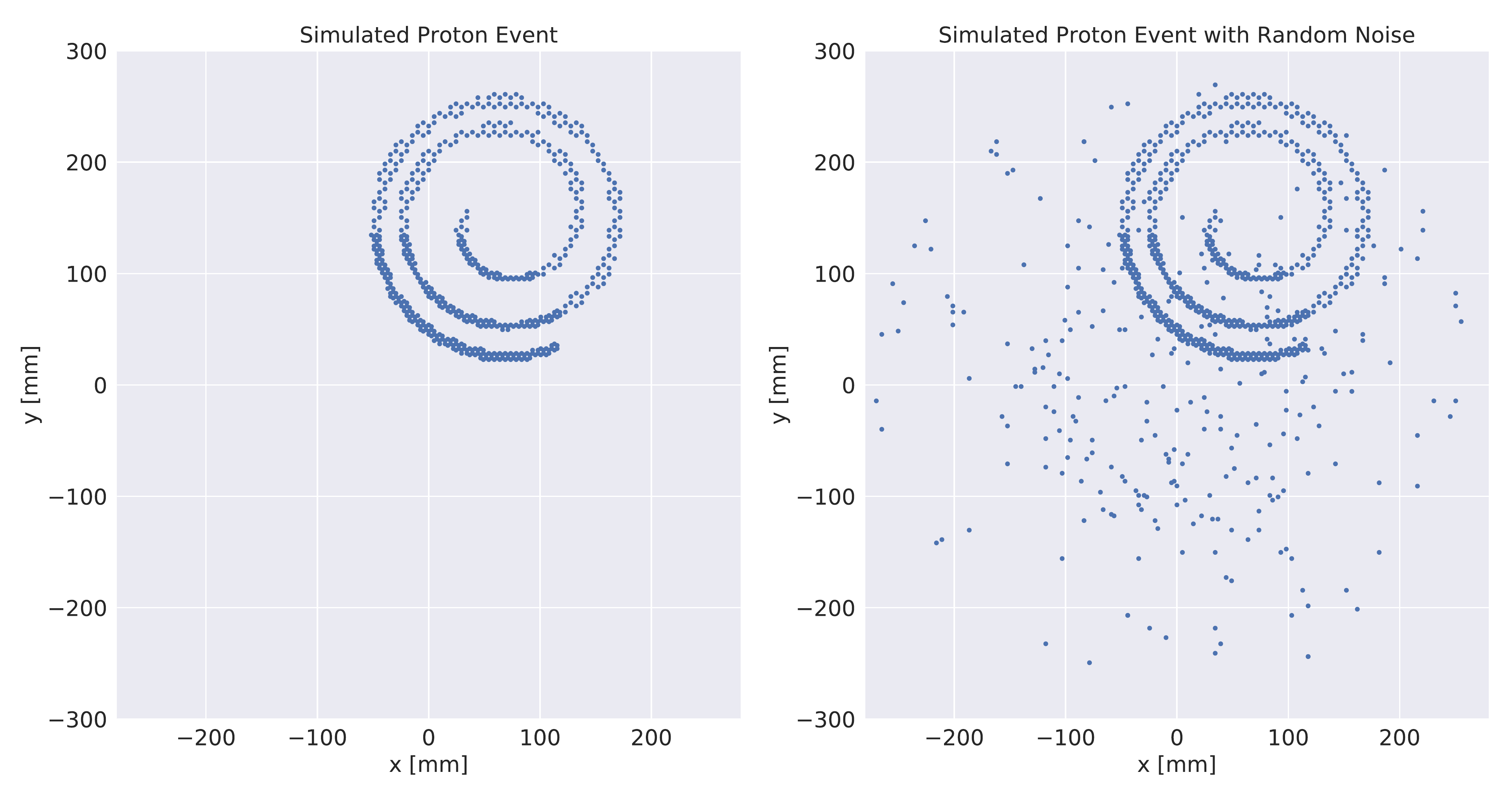}
\caption{The noise addition step for our simulated data. The left panel shows a clean simulated proton event while the right panel shows the same event with random noise added.}
\label{fig:add-noise}
\end{center}
\end{figure}

In our simulated proton and carbon events, we have a version of the data where we add random noise points to each track in hopes that our dataset more closely resembles the noisy data recorded in the highly-efficient AT-TPC. We add a random number of points at random $(x, y, z)$ coordinates within the detector chamber, each with a random charge value. Both the point count and the charge per point were generated from their own respective distributions based on statistics collected from data from the $^{46}Ar(p,p)$ experiment. Fig.~\ref{fig:add-noise} presents examples of our simulated events, with and without the presence of random noise.

\subsection{Detector volume discretization}
\begin{figure}[ht]
\begin{center}
\includegraphics[width=0.95\textwidth]{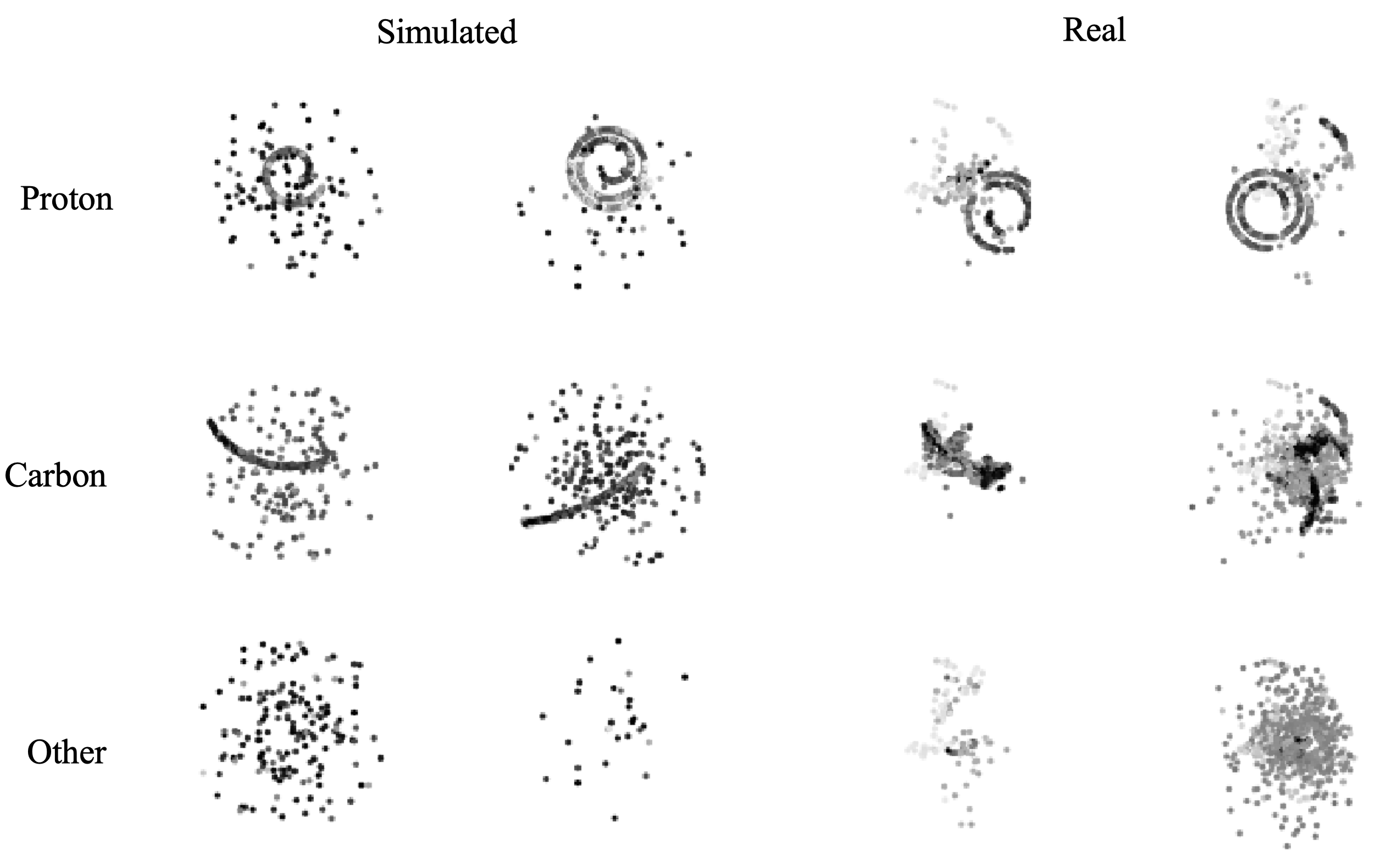}
\caption{A small subset of the training images for CNNs. The plots represent projections of the three-dimensional event data onto the $xy$-plane. The grayscale value indicates charge, where black is the highest charge and white represents no charge deposition. The simulated examples include added noise.}
\label{fig:CNN-data}
\end{center}
\end{figure}

Since our models require samples to be represented as fixed length feature vectors, we format our training data to capture the geometry of the detector. The detector pad plane has $10,240$ non-uniform pads that record data for $512$ time steps per event, producing a resolution of $5,242,880$ voxels in the detector volume. Training our models on a feature vector of this length is not tractable. For logistic regression and fully-connected neural networks, we decrease our resolution to $8000$ equal volume voxels by discretizing the $x$, $y$, and $z$ dimensions into 20 divisions each. Data was presented to the models as a one-dimensional vector of $8000$ elements, where each element's value represents the total charge recorded within that voxel's boundaries. For CNNs, we created a two-dimensional projection in $x$ and $y$ and discretized that projection into $16,384$ pixels, which can be represented by a $128 \times 128$ pixel image. In these images, the grayscale value is proportional to the total charge deposited onto the pads represented by each pixel. While one could also train 3D CNNs that can directly operate on the point-cloud data as collected in the detector, we choose to work with two-dimensional projections instead due to the easy availability of high-quality pre-trained 2D CNN models. Some examples of the data presented to the CNN models are shown in Fig.~\ref{fig:CNN-data}.

\subsection{Results}
\label{sec:results}

We now provide a brief overview of how the models described in Sec.~\ref{sec:ml} were implemented and tuned. We use the implementation of logistic regression from \texttt{scikit-learn}, an open-source machine learning library for Python \cite{scikit-learn}. This is an L2 regularized model, where the model parameters are determined using the SAG algorithm \cite{SAG}. SAG is a variant of stochastic gradient descent that guarantees faster convergence by retaining some memory of past gradient updates. The logistic regression results presented in this section report the performance obtained using the best value of the regularization constant $\lambda$, as found by a grid search on a test-by-test basis.\\

For our fully-connected neural network models, we use an architecture with a single hidden layer, implemented using Keras, a high-level deep learning library written in Python \cite{keras}. Keras is a wrapper for TensorFlow, a software library for streamlined numerical computation and automatic differentiation that is widely used in machine learning applications \cite{tensorflow}. The hidden layer in this model comprises 128 nodes and uses the ReLU activation function. We also employ dropout with a probability of $0.5$ to prevent overfitting. 
The output layer uses either the sigmoid or the softmax activation function, depending on whether the learning task is binary or multi-class. The network weights are learned using the Adam optimization algorithm \cite{adam}, a variant of stochastic gradient descent that uses an adaptive step size for each parameter that depends on the magnitude of recent gradient updates. We use a learning rate of $10^{-5}$ (tuned using a grid search) and employ early stopping to determine the optimal number of training iterations.\\

The CNN models used in this work build atop the \texttt{VGG16} model pre-trained on ImageNet \cite{vgg16}. As can be seen in Tab.~\ref{table:CNN_models}, we found that starting with the preset \texttt{VGG16} weights and fine-tuning the entire network using our datasets consistently outperformed using \texttt{VGG16} simply as a feature extractor (using the approach described in Sec.~\ref{sec:cnn}). Thus, all of our transfer learning based CNN tests in this section use the former approach. Separately, we also consider the performance of a CNN model trained from random initialization (i.e., without transfer learning) in Sec.~\ref{sec:simtoexp}. We replace the two fully-connected layers in the original \texttt{VGG16} architecture with a single, smaller fully-connected layer comprising $256$ units. We use dropout with a probability of $0.5$ when training this layer. Our output layer is composed of either two sigmoid units or three softmax units, depending on whether the learning task is binary or multi-class, in place of the original $1000$-unit output layer from \texttt{VGG16}. As with the fully-connected neural network models, we use Adam with a learning rate of $10^{-6}$ (determined using a grid search) to tune the model parameters and use early stopping to determine the number of training epochs. Finally, we note that all models and code created for this project are accessible through the associated GitHub repository \cite{ml-repo}.
\begin{table}[!htb]
	\centering
	\begin{tabular}{c|c| c|c}
	learning setting & learning task & \multicolumn{2}{c}{F1 score}\\  
	& & Feature extraction & Fine-tuning\\
	\hline
	\multirow{2}{*}{\exptoexp}  & binary & 0.89 & 0.90  \\
	& multiclass & 0.88 & 0.93 \\
	\hline
	\multirow{2}{*}{\simtoexp}  & binary & 0.67 & 0.72 \\
	& multiclass & 0.65 & 0.67\\ \hline
	\end{tabular}
	\caption{CNN model results: pre-trained CNN models were used as either fixed feature extractors, or were trained further starting from the preset weights. The resulting models were tested on experimental data. Fine-tuning the network outperformed feature extraction, and was thus the preferred transfer learning approach in other CNN tests.}
		\label{table:CNN_models}
\end{table}

\subsubsection{Training and testing on simulated data}

As noted earlier, training and testing on simulated data provides us with a useful upper-bound on how well we can expect our models to perform when they are evaluated on experimental data. We study the performance of three model families --- logistic regression (\textbf{LR}), fully-connected neural networks (\textbf{FCNN}) and fine-tuned convolutional neural networks that were pre-trained on ImageNet (\textbf{CNN}). Each of the algorithms were trained on $60,000$ events randomly selected from the overall set (see Tab.~\ref{table:datasets}), with $15\%$ of those events used for validation during training. The models were then evaluated on the remaining $24,000$ simulated events. Tab.~\ref{table:sim_results} presents the results from these tests. The results suggest that, given an ideal dataset, CNNs will perform best on the proton event classification task. In addition, CNNs display a robustness to the addition of statistical noise, which is not present in the other methods. 

\begin{table}[h!]
\centering

	\begin{tabular}{c| c |c| c | c| c}
	algorithm        & learning task & noise & precision & recall & F1 \\
\hline
	\multirow{2}{*}{LR}          &\multirow{2}{*}{binary}  & no & 0.98 & 0.99 & 0.98\\ & & yes &0.72 & 0.61 & 0.66\\ \cline{2-6}
    & \multirow{2}{*}{multiclass} & no & 0.98 & 0.99 & 0.98\\ & & yes & 0.77 & 0.69 & 0.73\\
    \hline
    \multirow{2}{*}{FCNN}          &\multirow{2}{*}{binary}  & no & 0.96 & 0.98 & 0.97\\ & & yes & 0.72 & 0.68 & 0.67\\  \cline{2-6}
    & \multirow{2}{*}{multiclass} & no & 0.96 & 0.98 & 0.97\\ & & yes & 0.76 & 0.72 & 0.74\\
    \hline
    \multirow{2}{*}{CNN}          &\multirow{2}{*}{binary}  & no & \textbf{1.00} & \textbf{1.00} & \textbf{1.00}\\ & & yes &  \textbf{1.00} & \textbf{1.00} & \textbf{1.00}\\  \cline{2-6}
    & \multirow{2}{*}{multiclass} & no & \textbf{1.00} & \textbf{1.00} & \textbf{1.00}\\ & & yes & 1.00 & 0.99 & 1.00\\
\hline
	\end{tabular}

\caption{Results in the \simtosim~regime: simulated data was used for training and testing each of our algorithms. The simulated data was also augmented with noise and retested to better understand how our model performance could be affected when applied to noisy events from the experimental data.}
\label{table:sim_results}
\end{table}

\subsubsection{Training and testing on experimental data}

Training and testing the models on experimental data requires manually labeling this data, which is not desirable in the analysis workflow. Further, human annotation of training data, while relatively straightforward in our classification task, is not always as simple. 
However, the results on this task help us identify the algorithms that succeed with our experimental data, which has noise signatures and events that we are unable to simulate in our synthetic datasets. We randomly selected $2151$ events (of the $2689$ hand-labeled events shown in Tab.~\ref{table:datasets}) to create our training set, with $15\%$ of those events used for validation during training, and reserved the other $538$ events for evaluating the model. The results from these tests are presented in Tab.~\ref{table:real_results}, and, like the simulated data results in Tab.~\ref{table:sim_results}, suggest that the best algorithm for proton event classification is a pre-trained CNN fine-tuned on our data. While the workflow is not ideal, training on experimental data provides us with a classification model that is sufficiently successful to use in the analysis pipeline. \\

\begin{table}[htb]
    \centering
	\begin{tabular}{c| c |c| c | c}
	algorithm        & learning task & precision & recall & F1 \\
    \hline
	\multirow{2}{*}{LR}          &binary  & 0.78 & 0.58 & 0.66 \\ \cline{2-5}
    & multiclass &  0.77 & 0.67 & 0.72\\
    \hline
    \multirow{2}{*}{FCNN}          &binary  & 0.85 & 0.54 & 0.66 \\ \cline{2-5}
    & multiclass & 0.83 & 0.62 & 0.71\\
    \hline
    \multirow{2}{*}{CNN}          &binary  & 0.98 & 0.84 & 0.90 \\ \cline{2-5}
    & multiclass &  \textbf{0.96} & \textbf{0.90} & \textbf{0.93}\\
    \hline
	\end{tabular}
	\caption{Results in the \exptoexp~regime: these results were obtained by training and testing on hand-labeled data. The train and test sets came from a small dataset of fewer than 3000 events as reported in Tab.~\ref{table:datasets}.}
    \label{table:real_results}
\end{table}

\begin{figure}[htb]
    \centering
    \includegraphics[width=0.85\textwidth]{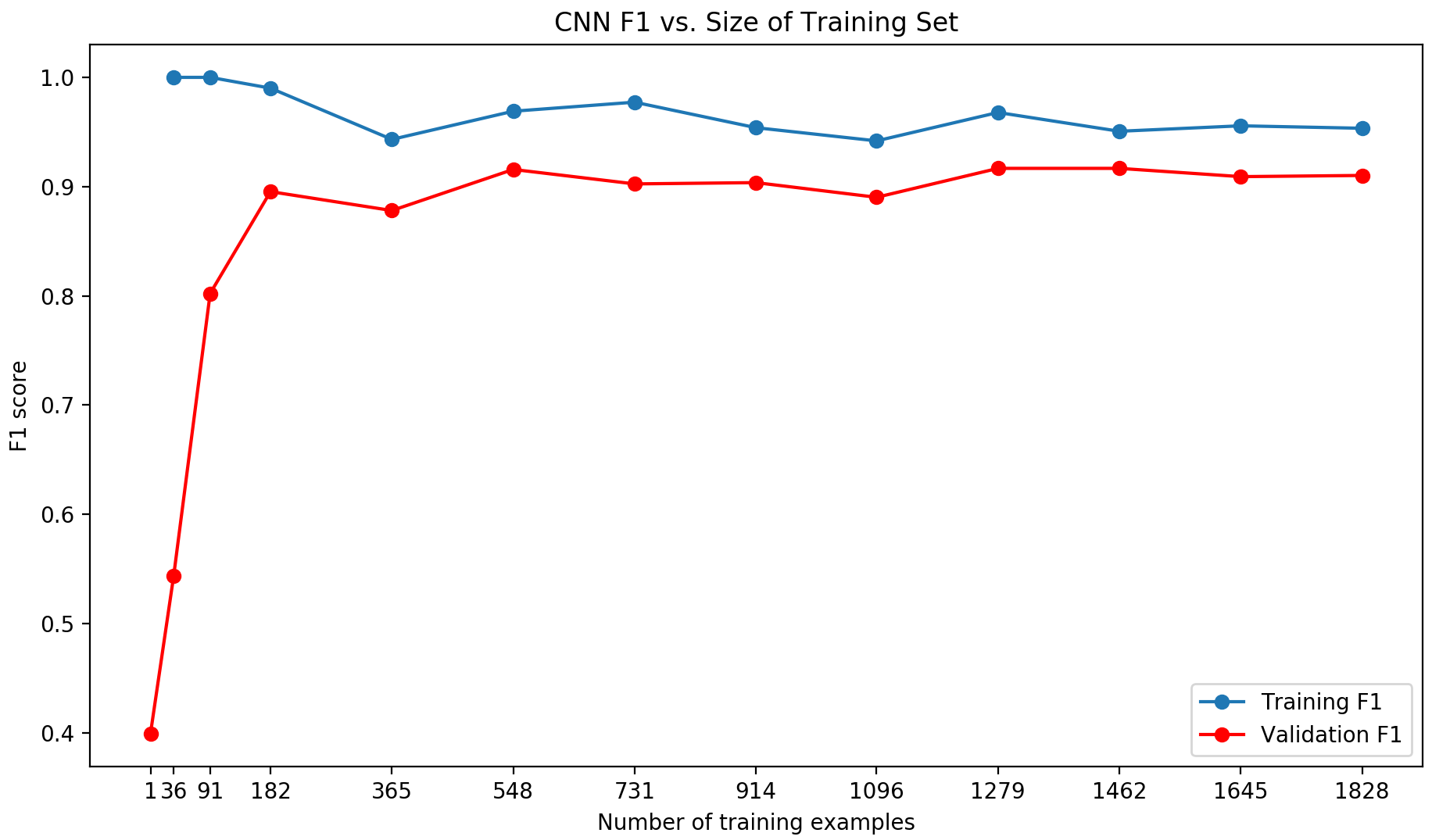}
    \caption{CNN performance as a function of the number of examples used for training. The training sets are constructed in a cumulative fashion, so that each larger training set fully contains the smaller ones. The red validation curve is generated by evaluating the trained model on a fixed set of $323$ examples ($15\%$ of the total training data).}
    \label{fig:learning-curve-bias-variance}
\end{figure}

Intriguingly, we also find that CNNs are sample efficient on this learning task. Fig.~\ref{fig:learning-curve-bias-variance} presents the performance of our CNN as we increase the size of the training set used to fine-tune the initial ImageNet-based weights of the model. We find that the performance of the model reaches its peak after training on only {\small$\sim$}$550$ examples. This suggests that at least in some experimental settings such as ours, where humans can annotate data reliably, large-scale labeling of data may not be necessary for creating high-performing classifiers.

\subsubsection{Training on simulated data and testing on experimental data}
\label{sec:simtoexp}

The goal for this work is to successfully classify experimental data after training a model on simulated data. This is ideal from an experimentalist's perspective, as this frees them from having to manually label any data after the experiment has run. This approach also has the advantage of being applicable in situations where human labeling of data is challenging or unreliable. In this task, we used the same training set of $60,000$ simulated events as from the \simtosim~tests, with $15\%$ of these examples used for validation. The full set of $2,689$ labeled experimental events shown in Tab.~\ref{table:datasets} was used for model evaluation. Tab.~\ref{table:transfer-results} reports our results in this \simtoexp~setting. As expected given our simulated and experimental results, fine-tuned CNNs performed the best at classifying proton scattering events in the experimental data, though the F1 scores have decreased significantly from the previous tests. This is expected since our simulated data does not perfectly model our experimental data.\\

\begin{table}[htb]
    \centering
	\begin{tabular}{c| c |c| c | c| c}
	algorithm        & learning task & noise & precision & recall & F1 \\
    \hline
	\multirow{2}{*}{LR}          &\multirow{2}{*}{binary}  & no & 0.41 & 0.36 & 0.39\\ & & yes & 0.36 & 0.32 & 0.34\\ \cline{2-6}
    & \multirow{2}{*}{multiclass} & no & 0.44 & 0.45 & 0.45\\ & & yes &0.35 & 0.28 & 0.31\\
    \hline
    \multirow{2}{*}{FCNN}          &\multirow{2}{*}{binary}  & no & 0.37 & 0.41 & 0.39\\ & & yes & 0.34 & 0.32 & 0.33\\  \cline{2-6}
    & \multirow{2}{*}{multiclass} & no & 0.38 & 0.53 & 0.44\\ & & yes & 0.35 & 0.30 & 0.32\\
    \hline
    \multirow{2}{*}{CNN}          &\multirow{2}{*}{binary}  & no & \textbf{0.90} & \textbf{0.60} & \textbf{0.72}\\ & & yes & 0.39 &0.91  & 0.55\\  \cline{2-6}
    & \multirow{2}{*}{multiclass} & no & 0.96 & 0.52 & 0.67\\ & & yes & 0.27 & 0.99 & 0.42\\
    \hline
	\end{tabular}
    \caption{Results in the \simtoexp~regime: this is a more difficult learning problem, since the training and test data come from different distributions. Nevertheless, we see successful classification of proton scattering events using CNNs.}
    \label{table:transfer-results}
\end{table}

Finally, we consider one further \simtoexp~setup, where we train a model with the \texttt{VGG16} architecture, but starting from a random initialization of the network weights. We did not consider this test in the \simtosim~setting, since the CNNs achieved high performance in that scenario with transfer learning techniques. The test was infeasible in the \exptoexp~setting, given the small size of the labeled experimental dataset. In this \simtoexp~setting on the other hand, we can investigate whether training a CNN from scratch helps us gain a performance edge over starting from pre-trained weights. We train a CNN on a large dataset of $10^6$ simulated events with binary class labels, and use a further $200,000$ examples for validation. We then attempt to classify the $2689$ labeled experimental events from Tab.~\ref{table:datasets} using this model (this approach is similar to that proposed in \cite{Adams:2018bvi}). In a further test, we fine-tune this trained model with a subset of the labeled experimental data ($2151$ events), and evaluate the performance of this model on the remaining $538$ events (mirroring  the approach described in \cite{transfer-quantumchem}). The results from these tests are presented in Tab.~\ref{table:retrain-cnn-results}. We note that training purely on the simulated data only yields an F1 score of $0.44$ on the experimental data, which is significantly lower than the CNN models that employed transfer learning and were presented in Tab.~\ref{table:transfer-results}. Fine-tuning these models on experimental data improves the performance, though these models still fall short of the CNN models used in Tab.~\ref{table:real_results}. Given that training these CNNs afresh required almost $15$ hours of GPU time (compared to just minutes for the pre-trained models) and resulted in no performance gains, we recommend that pre-trained models and transfer learning be preferred when possible.

\begin{table}[htb]
    \centering
	\begin{tabular}{c|c|c|c|c}
	training process & noise & precision & recall & F1 \\ \hline
	\multirow{2}{*}{without tuning} & no & 0.30 & 0.83 & 0.44 \\
	& yes & 0.39 & 0.44 & 0.41 \\ \hline
	\multirow{2}{*}{with tuning} & no & 0.90 & 0.92 & 0.91 \\
	& yes & 0.93 & 0.85 & 0.89 \\
    \hline
	\end{tabular}
    \caption{Results in the \simtoexp~regime without transfer learning: the \texttt{VGG16} architecture was initialized with random weights and then trained on a large set of simulated events using binary class labels. These weights are then optionally tuned further on a subset of the labeled experimental data. The final model is evaluated on held out experimental data.}
    \label{table:retrain-cnn-results}
\end{table}

\subsubsection{Visualizing the learning results}

\begin{figure}[htb]
    \centering
    \begin{tabular}{c!{\vrule width 1pt}c}
    \includegraphics[width=0.23\textwidth]{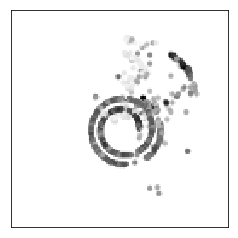}
    \includegraphics[width=0.23\textwidth]{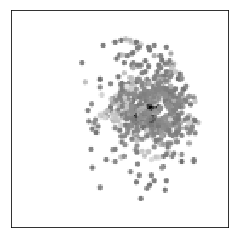}
    &
    \includegraphics[width=0.23\textwidth]{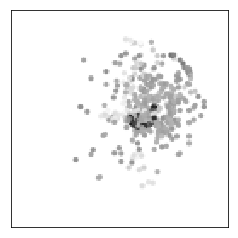} 
    \includegraphics[width=0.23\textwidth]{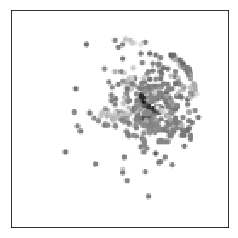}\\
    \includegraphics[width=0.23\textwidth]{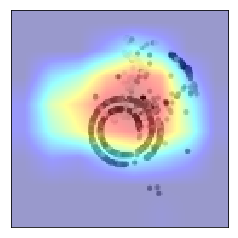}
    \includegraphics[width=0.23\textwidth]{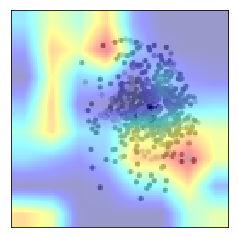}
    &
    \includegraphics[width=0.23\textwidth]{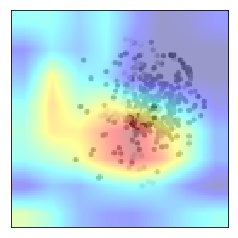}
    \includegraphics[width=0.23\textwidth]{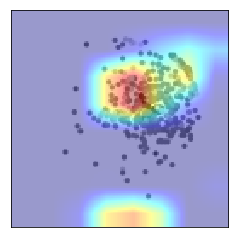}
    \end{tabular}
    \caption{Sample visual explanations of the CNN's classification decisions on experimental data. The top row shows the input images. The heatmaps on the bottom row indicate the regions of the respective images that the model paid particular attention to when making its classification decisions. Areas shaded in red correspond to pixels that were weighted more heavily in the decision, while areas shaded in blue were weighted less. From left-to-right, these examples constitute cases where the model correctly labeled a proton event, correctly labeled a non-proton event, mistook a proton event for a non-proton event, and mistook a non-proton event for a proton event. We note that the model focuses on regions of the point cloud that display structure when correctly identifying proton events. Its ``attention'' is more diffuse in the other scenarios.}
    \label{fig:heatmaps}
\end{figure}

In this section, we interrogate our fitted CNN model (pre-trained on ImageNet and then fine-tuned on experimental data) to better understand how classification decisions are made. Specifically, we construct visual explanations of the model's predictions using the Gradient-weighted Class Activation Mapping (Grad-CAM) algorithm \cite{grad-cam}. Four representative examples are shown in Fig.~\ref{fig:heatmaps}. The top row in the figure shows the input images, while the heatmaps in the bottom row highlight the degree to which each region in an image contributed to the model's overall decisions. The left-half of Fig.~\ref{fig:heatmaps} presents cases where the model made correct predictions, on a proton event and an unknown event (arranged left-to-right). We see that the model's ``justification'' for the proton classification arises from its focus on the spiral structure in that image, while its attention is more dispersed in the case of the unknown event. In the right-half of Fig.~\ref{fig:heatmaps}, we see cases where the model made incorrect predictions, namely a proton event mistakenly classified as non-proton, and a non-proton event incorrectly classified as a proton (again arranged left-to-right). The model's errors in these cases are understandable given the challenging identification task posed by these examples.

%% file: conclusions.tex
\section{Conclusions}

We found that machine learning methods are effective at identifying proton events in both simulated and real AT-TPC data. Most significantly for experimentalists, we were able to construct models that could learn from artificial data and apply this knowledge in classifying actual data collected from the detector. We also demonstrated the effectiveness of transfer learning in this problem domain, by fine-tuning pre-trained CNNs. The ability to train a model prior to conducting the physics experiment would allow automatic classification of events nearly in real-time, with minimal need for human supervision. Further, such an approach could be easily adapted to other experiments, that utilize different instruments or detector technology, as all that is required is access to an accurate physics simulator.\\

Our highest performing model --- a pre-trained convolutional neural network that was then fine-tuned for our domain --- achieved an impressive precision of $0.90$ on the transfer learning task. We note, however, that there is still room for improvement, for this same model only has a recall of $0.60$, for an F1 score of $0.72$. In comparison, our best performing CNN model that also trained on experimental data was able to achieve an F1 score of $0.93$. However, the latter approach comes at a cost: it requires laborious hand-labeling of experimental data, before the classifier can be built. This approach may fail in other experiments where hand-labeling the data is error-prone or infeasible. Our results indicate that augmenting our simulated data with statistical noise is not a sufficient proxy to assist in our classification. We believe that the performance gap can be closed with simulations that better capture the structural noise that is observed in experimental data --- an avenue that we plan to pursue in future work.

%% file: acknowledgements.tex
\section{Acknowledgements}
This work supported in part by NSF grant no.~10049216 and the Davidson College Faculty Study and Research Grant.